\documentclass[runningheads]{llncs}

\usepackage{style}

\usepackage{styleabbrv}

\usepackage{graphicx}
\usepackage{booktabs}

\usepackage[accsupp]{axessibility}  % Improves PDF readability for those with disabilities.

\usepackage[pagebackref,breaklinks,colorlinks,citecolor=styleblue]{hyperref}

\usepackage{orcidlink}

\usepackage{wrapfig}
\usepackage{multirow}

\usepackage{newfloat}

\usepackage{times}
\usepackage{epsfig}
\usepackage{graphicx}
\usepackage{amsmath}
\usepackage{amssymb}
\usepackage{pifont}
\usepackage{booktabs}
\usepackage{arydshln}

\usepackage{colortbl}
\usepackage{caption}
\usepackage{subcaption}
\usepackage{multirow}
\usepackage{rotfloat}
\usepackage{capt-of}
\usepackage{longtable}
\usepackage{diagbox}
\usepackage{makecell}
\usepackage{enumitem}

\newcommand{\cmark}{\ding{51}}%
\newcommand{\xmark}{\ding{55}}%

\usepackage{subcaption} % Load the subcaption package
\usepackage{lipsum} % Provides sample text
\usepackage{tabularx} % Allows for table columns to be automatically sized

\makeatletter
\def\@fnsymbol#1{\ensuremath{\ifcase#1\or \dagger\or *\or \ddagger\or
   \mathsection\or \mathparagraph\or \|\or \dagger\dagger
   \or **\or \ddagger\ddagger \else\@ctrerr\fi}}
\makeatother
\newcommand*\samethanks[1][\value{footnote}]{\footnotemark[#1]}

\begin{document}

\title{DriveCoT: Integrating Chain-of-Thought Reasoning with End-to-End Driving} 

\titlerunning{DriveCoT}

\author{Tianqi Wang\inst{1}\orcidlink{0009-0000-5636-8013} \and
Enze Xie\inst{2}\thanks{Corresponding authors.}\orcidlink{0000-0001-6890-1049} \and
Ruihang Chu\inst{3}\orcidlink{0000-0001-9057-745X} \and Zhenguo Li\inst{2} \and Ping Luo\inst{1}\samethanks\orcidlink{0000-0002-6685-7950}}

\authorrunning{Tianqi Wang et al.}

\institute{The University of Hong Kong \and
Huawei Noah’s Ark Lab \and
The Chinese University of Hong Kong\\
\email{wangtq@connect.hku.hk, johnny\_ez@163.com, \\ rhchu@cse.cuhk.edu.hk, li.zhenguo@huawei.com, pluo@cs.hku.hk}}

\maketitle

\begin{abstract}
End-to-end driving has made significant progress in recent years, demonstrating benefits such as system simplicity and competitive driving performance under both open-loop and closed-loop settings.
Nevertheless, the lack of interpretability and controllability in its driving decisions hinders real-world deployment for end-to-end driving systems. In this paper, we collect a comprehensive end-to-end driving dataset named DriveCoT, leveraging the CARLA simulator. It contains sensor data, control decisions, and chain-of-thought labels to indicate the reasoning process. We utilize the challenging driving scenarios from the CARLA leaderboard 2.0, which involve high-speed driving and lane-changing, and propose a rule-based expert policy to control the vehicle and generate ground truth labels for its reasoning process across different driving aspects and the final decisions. This dataset can serve as an open-loop end-to-end driving benchmark, enabling the evaluation of accuracy in various chain-of-thought aspects and the final decision. In addition, we propose a baseline model called DriveCoT-Agent, trained on our dataset, to generate chain-of-thought predictions and final decisions. The trained model exhibits strong performance in both open-loop and closed-loop evaluations, demonstrating the effectiveness of our proposed dataset. Project page: \url{https://drivecot.github.io/}.

\end{abstract}

\section{Introduction}
\label{sec:intro}
In recent years, autonomous driving has gained significant development owing to the progress made by both academia and industry \cite{survey1, survey2, survey3}. Currently, the main-stream technical solutions for autonomous driving adopt modular designs, consisting of perception, prediction, and planning \& control with hand-crafted rules to connect different modules. This is due to the interpretability and controllability brought by the modular design to ensure driving safety. Nevertheless, the modular design has some fundamental limits, such as the complex system design and the error accumulation across modules. Another research direction raising in recent years is end-to-end driving, where the system accepts sensor data and generates planning results \cite{transfuser, hiddenbias, TCP, interfuser, reasonnet, thinktwice, neat_high_level} or even low-level control signals directly \cite{improvedIL, e2e_low_level1, e2e_low_level2, e2e_low_level3, interpretable_e2e_low_level_rl, roach_low_level_rl, low_level_rl_new}. This kind of approaches have great system simplicity and have shown competitive driving performance. However, the lack of explainability and controllability caused by end-to-end methods' black-box nature hinders their deployment in real-world. 

\begin{figure}[tb]
  \centering
  \includegraphics[width=1.0\linewidth]{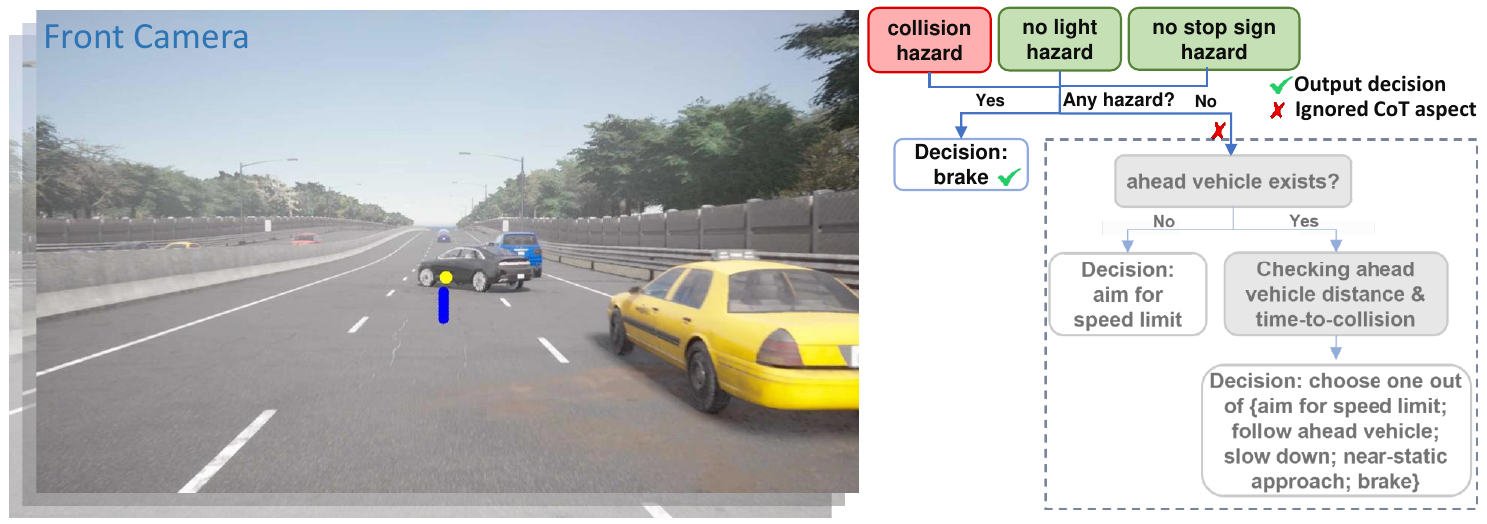}
  \caption{
 We propose DriveCoT, which includes a new dataset, benchmark, and baseline model for end-to-end autonomous driving. Sensor data such as camera images along with a direction-indicating target point (yellow dot in the left image) are given as the model inputs. As shown in the right part, the model obtains the final speed decision by generating predictions on different driving aspects and 
  conducting chain-of-thought reasoning. In addition, our model generates the planned future waypoints (blue dots in the left image) for steering.}
  \label{fig:dataset_example}
\end{figure}

To mitigate the black-box issues for end-to-end driving, increasing number of recent works focus on the explainability and the reasoning process when making driving decisions \cite{interpretable_e2e_1, interpretable_e2e_2, interpretable_e2e_3, nuscenesQA, nuprompt, bdd-x, drivegpt4, drama, rank2tell, lmdrive, drivemlm, drivelm}. These studies proposes large-scale driving dataset with sensor data and QA pairs for one or several modules, including perception, prediction, and planning either relying on existing datasets with additional QA annotations or collecting new datasets. The QA pairs provided by these studies either annotate the relevant objects that can affect driving decisions or summarize the reasons for specific driving actions, thus enhancing the explainability. However, most of these datasets are built on existing datasets such as nuScenes \cite{nuscenes} and BDD \cite{bdd},  which contain only normal and safe driving scenarios and provide single explanation for the driving decisions. Notably, DriveMLM \cite{drivemlm} utilizes the challenging driving scenarios from the CARLA leaderboard 2.0 benchmark to collect the driving data. Yet, it only provides the driving decision with single explanation without the derivation process, which limits the interpretability. Another recent work DriveLM \cite{drivelm} annotates the reasoning process of the driving decision as a Graph Visual Question Answering (GVQA), where the QAs are interconnected and need multi-round model inferences, resulting in extremely long inference time. Moreover, no challenging and safety-critical driving scenarios are involved in DriveLM.

In this paper, we propose a new dataset named DriveCoT, which includes challenging driving scenarios such as high-speed driving and lane-changing. 
Besides, we annotate the driving decisions with the chain-of-thought labels for different driving aspects to indicate the detailed thinking process, as shown in Fig.~\ref{fig:dataset_example}. These labels are automatically generated by our proposed rule-based expert policy, which utilizes the ground truth states of the CARLA simulator.
The rule-based expert policy is utilized to control the ego vehicle to handle the challenging scenarios proposed in CARLA leaderboard 2.0 \cite{leaderboard2.0}. Various sensors, including multi-view cameras and lidar, are attached to the ego vehicle to collect sensor data for the DriveCoT dataset. Furthermore, we store the expert's chain-of-thought results along with the final decisions as ground truth labels to offer interpretability for the decision-making process of end-to-end driving.

In addition, we propose a baseline model called DriveCoT-Agent. It takes the multi-view camera video for a past period and a direction-indicating target point as inputs to generate the chain-of-thought (CoT) predictions on different driving aspects and planned future waypoints. Instead of using single frame images only as input such as the methods in \cite{interfuser, transfuser, hiddenbias, drivelm}, we leverage the multi-view camera video which captures the motion of both ego vehicle and surrounding objects, thus enabling early prediction of potential hazards and supporting high-speed driving. Additionally, the final driving decisions can be derived from the model's chain-of-thought predictions via a process demonstrated in Fig.~\ref{fig:CoT process}. Besides interpretability, the trained model outperforms previous methods in both open-loop evaluation on DriveCoT validation data and closed-loop testing benchmarks by significant margins.

Our main contributions can be summarized as follows:
\begin{itemize}
    \item We introduce DriveCoT, the first end-to-end driving dataset containing chain-of-thought thinking process labels and diverse challenging driving scenarios.
    \item We design a new rule-based expert policy to handle the challenging scenarios in CARLA leaderboard 2.0, thus controlling the ego vehicle effectively and storing the thinking process labels.
    \item We propose a novel baseline model termed DriveCoT-Agent. It is trained on DriveCoT dataset to generate the chain-of-thought predictions and the final driving decisions, showing strong performance on both open-loop and closed-loop evaluations. 
\end{itemize}

\section{Related work}

\subsection{End-to-End Autonomous Driving}
With the open-sourced driving simulator CARLA released in recent years, a series of works regarding end-to-end driving have been published \cite{transfuser, hiddenbias, TCP, interfuser, reasonnet, thinktwice, neat_high_level, improvedIL, e2e_low_level1, e2e_low_level2, e2e_low_level3, interpretable_e2e_low_level_rl, roach_low_level_rl, low_level_rl_new}.
These methods generally share a similar pipeline, using the target point and sensor data as inputs to produce high-level planned waypoints or low-level control commands.
To further enhance interpretability, existing works \cite{interpretable_e2e_1, interpretable_e2e_2, interpretable_e2e_3, interfuser} develop auxiliary tasks with semantic meanings such as 3D object detection and road segmentation.
Specifically, InterFuser \cite{interfuser} generates intermediate predictions towards traffic signals and object occupancy map, which are then leveraged by downstream controller to explicitly enhance driving safety.
Besides, Transfuser++ \cite{hiddenbias} studies on the common yet ignored issues in existing works, finding that separated and disentangled predictions for target speed and planned waypoints can offer flexibility and improve driving performance. Moreover, it finds that the target point condition is beneficial for the steering recovery compared to discrete navigation commands.

\begin{table}[t!]
    \caption{Comparison of AD datasets for driving understanding. DriveCoT offers labels for all aspects with chain-of-thought connections and contains diverse challenging driving scenarios.}
    \vspace{-1mm}
    \begin{center}
    \renewcommand{\tabcolsep}{1mm}
        \resizebox{0.85\linewidth}{!}{
        \begin{tabular}{l|c c c c|c}
            \toprule[0.3mm]
            Dataset & Perception & Prediction & Planning & Reasoning & Challenging Scenarios \\
            \midrule
            nuPrompt \cite{nuprompt} & \cmark & \xmark & \xmark & none & \xmark\\
            nuScenes-QA \cite{nuscenesQA} & \cmark & \xmark & \xmark & single & \xmark\\
            Rank2Tell \cite{rank2tell} & \cmark & \xmark & \xmark & single & \xmark\\
            DRAMA \cite{drama} & \xmark & \xmark & \cmark & single & \xmark\\
            BDD-X \cite{bdd-x} & \xmark & \xmark & \cmark & single & \xmark\\
            DriveGPT4 \cite{drivegpt4} & \xmark & \xmark & \cmark & single & \xmark\\
            LMDrive \cite{lmdrive} & \xmark & \xmark & \xmark & none & \cmark\\
            DriveMLM \cite{drivemlm} & \cmark & \xmark & \cmark & single & \cmark\\
            DriveLM \cite{drivelm} & \cmark & \cmark & \cmark & graph & \xmark\\
            \textbf{DriveCoT ({ours})} & \cmark & \cmark & \cmark & CoT & \cmark\\
            \bottomrule[0.3mm]
        \end{tabular}
        }
    \label{tab:dataset_comparison}
    \end{center}
     \vspace{-4mm}
\end{table}
% \vspace{-10mm}
Our proposed DriveCoT-Agent adapts the positive findings in previous works, including the interpretable intermediate predictions and disentangled target speed from planned waypoints. In addition, we employ video input from multi-view cameras rather than single-frame images, enabling the encoding of object motions. While existing works mainly focus on driving safety, thereby reducing the driving decisions to either normal driving or braking, the proposed DriveCoT-Agent further considers comfort and alignment with surrounding traffic. It can generate more nuanced speed decisions, such as following ahead vehicle or slowing down.

\subsection{Driving with Understanding}

To bridge the gap for real-world deployment, an arising branch in end-to-end driving focuses on driving with understanding \cite{nuscenesQA, nuprompt, bdd-x, drivegpt4, drama, rank2tell, lmdrive, drivemlm, drivelm}. 
These works often introduce large-scale driving datasets with scene understanding annotations either based on existing datasets such as nuScenes \cite{nuscenes} and BDD \cite{bdd} or collecting new datasets, with some including challenging driving scenarios. The scene understanding annotations span over the perception, prediction, and planning modules of driving system.  Existing works such as nuPrompt \cite{nuprompt}, nuScenes-QA \cite{nuscenesQA}, and Ran2Tell \cite{rank2tell} offers object-level QAs for perception module.
DRAMA \cite{drama}, BDD-X \cite{bdd-x}, and DriveGPT4 \cite{drivegpt4} offers planning results with single explanation. LMDrive \cite{lmdrive} and DriveMLM \cite{drivemlm} incorporate challenging driving scenarios via the usage of the CARLA simulator yet only have none or a single explanation for reasoning. 
DriveLM \cite{drivelm} is a recently published dataset which contains driving understanding QAs across different modules. However, DriveLM contains only normal driving scenarios from nuScenes and CARLA.

As illustrated in Table~\ref{tab:dataset_comparison}, our proposed DriveCoT dataset consists of the most comprehensive aspects for driving understanding, integrated with CoT relations, and covers a wide range of challenging driving scenarios.

\subsection{Chain-of-Thought Reasoning}
Chain-of-thought reasoning is a sequential process where a complicated task is broken down into a series of easier logical steps to reduce the complexity and enhance the performance. This kind of reasoning is similar to how humans approach complicated tasks with intermediate steps. Chain-of-thought reasoning has applications in explainable AI (XAI) \cite{cot_related1, cot_related2, cot_related3} to offer explanations for network predictions, which can be crucial for safety-related tasks such as driving or medical applications \cite{interpretable_e2e_low_level_rl, xai_medical}. With the rise of the Large Language Model (LLM), chain-of-thought reasoning has also shown performance benefits in both general commonsense tasks \cite{cot_general_llm1, cot_general_llm2} and more application-orientated tasks, including robotics \cite{cot_robotics} and autonomous driving \cite{cot_drivelikehuman, cot_mpc}. Specifically, \cite{cot_drivelikehuman} and \cite{cot_mpc} purely focus on the decision-making module of autonomous driving and design chain-of-thought frameworks to leverage LLM's reasoning ability. Similar to our settings, DriveLM \cite{drivelm} aims at end-to-end driving and takes sensor data as inputs, requiring the ability for perception, prediction, and decision-making. Nevertheless, its proposed baseline model requires multi-round inference of LLM due to its dynamic graph-structured VQA pairs, resulting in extremely long inference time.

We adapt chain-of-thought reasoning to decompose the driving task into several representative subtasks (see Fig.~\ref{fig:CoT process}), including collision prediction, traffic sign recognition, relation to ahead vehicle, and road recognition. Given the real-time inference requirements for real-world driving deployment, we propose a baseline model named DriveCoT-Agent that achieves satisfactory performance with fast inference, whereas the LLM counterparts \cite{drivelm, drivemlm, cot_drivelikehuman, cot_mpc} take at least several seconds for one step.

\section{DriveCoT Dataset}
\subsection{Data Collection Process}
We collect data using CARLA 0.9.14 version under the leaderboard 2.0 framework \cite{leaderboard2.0} with our proposed rule-based
expert policy modified from \cite{carlaexpert} to adapt to high-speed driving and more challenging scenarios. 
Additionally, we use a set of predefined routes across urban, residential, rural, and highway areas to execute the expert policy and drive the ego vehicle while encountering numerous challenging scenarios.  
For each scenario, data collection is initiated at a predefined trigger point and stopped upon either surpassing 20 seconds of simulated time or reaching the trigger point for the next scenario.

\begin{figure}[tb]
  \centering
  \includegraphics[width=1.0\linewidth]{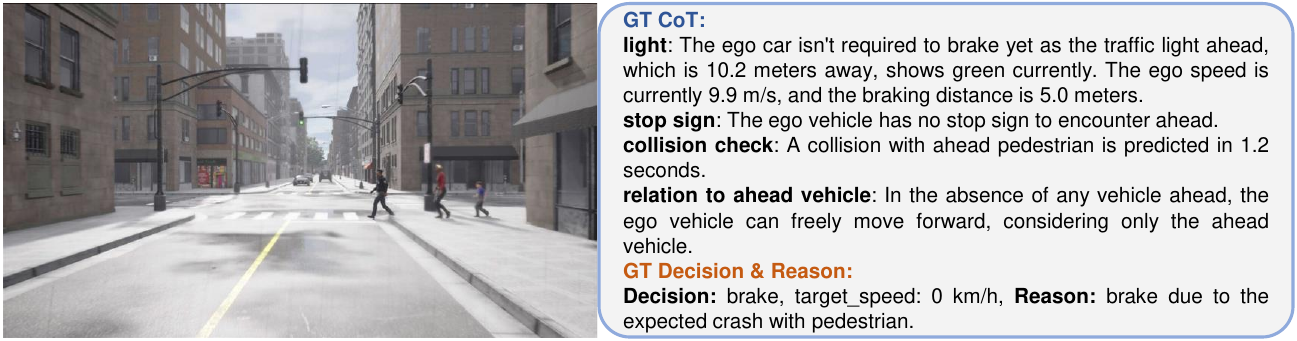}
  \caption{Text-form annotations in our proposed DriveCoT dataset. Descriptions for each chain-of-thought aspects and the final decision \& reason are rewritten and diversified using ChatGPT.
  }
  \label{fig:text annotation}
\end{figure}

During data generation, we collect sensor data, including images from multi-view cameras and lidar point clouds from ego vehicle, and record the expert policy's reasoning process and decisions as driving understanding labels in a chain-of-thought format (see Fig. \ref{fig:CoT process}). For ground-truth labels, we offer both text-form annotations (Fig. \ref{fig:text annotation}) and simplified classification results (Fig. \ref{fig:stats_CoT}) for flexible usage. The text annotations are diversified by ChatGPT to vary sentence structures for each CoT aspect, also ensuring accurate insertion of specific details like distance and speed.

The DriveCoT dataset comprises 1058 scenarios and 36K labeled samples (similar to nuScenes), collected at a 2 Hz frequency, averaging 17 seconds per scenario. 
We partition the dataset into training, validation, and testing sets at a ratio of 70\%, 15\%, and 15\%, respectively, yielding 25.3K training, 5.5K validation, and 5.5K testing samples. To prevent data leakage, we assign all data from the same scenario to the same set. Additionally, we ensure that the distribution of CoT aspects is similar across all splits.

\begin{figure}[tb]
  \centering
  \includegraphics[width=0.8\linewidth]{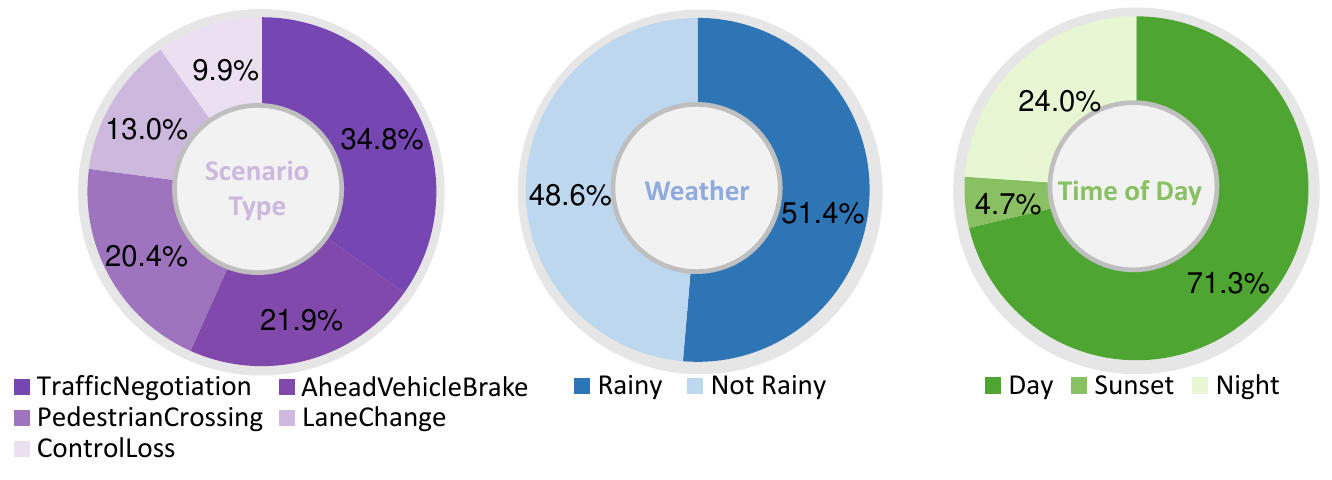}
  % \vspace{-2mm}
  \caption{Scenario-level distribution over scenario type, weather, and time of day. DriveCoT consists of various types of challenging driving scenarios and diverse weather and time-of-day conditions.
  }
  \label{fig:stats_scene}
\end{figure}

\subsection{Expert Policy}
We propose a rule-based expert policy that has access to ground truth states in the simulator, which mainly follows the method in \cite{carlaexpert}. With effective modifications, we make it work for high-speed driving in leaderboard 2.0. We design dynamic braking distance for the ego vehicle based on ego speed to detect potential hazards, including red traffic lights, stop signs, or surrounding vehicles and pedestrians. Besides, the proposed expert policy also considers the relation to the ahead vehicle in the same lane to generate more nuanced speed decisions. For the planned future waypoints, we collect expert waypoints with fixed distance-gap, similar to Transfuser++ \cite{hiddenbias}, instead of a fixed time-gap to disentangle waypoints from the target speed. Also, the planned waypoints are chosen further from the ego vehicle when the ego speed increases to avoid oscillating. More details can be found in the supplementary material.

In DriveCoT, we organize the collected data based on scenarios. 
Each scenario has a meta file indicating the scenario type, weather condition, and time of day.
Each frame sample can be correlated to a specific scenario according to the file name. 
Each frame contains the sensor data from six 1600$\times$900 RGB cameras and a 32-lane LiDAR sensor, along with the expert policy's decision-making process labels and final decisions in both text form and simplified classification form. As shown in Fig.~\ref{fig:CoT process}, the CoT aspects include checking red traffic light hazard, stop sign hazard, potential collision with surrounding objects, relation to ahead vehicle, and etc.

\subsection{Dataset Statistics}

In this section, we show the statistics of DriveCoT in terms of the ego vehicle speed, scenario-level
attributes, and sample-level CoT annotations \& final decisions.

\subsection{Dataset Format}

\setlength\intextsep{-0pt}
\begin{wrapfigure}{r}{6cm}
    \centering
    \includegraphics[width=0.4\textwidth]{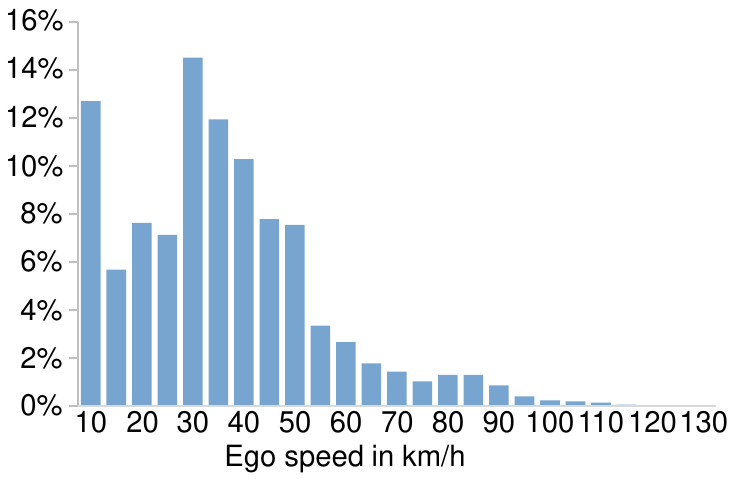}
    \caption{Distribution of ego speed in DriveCoT dataset, which includes a significant amount of high-speed driving data (above 60 km/h).}
    \label{fig:ego_speed}
\end{wrapfigure}

\textbf{Ego speed.} 
Existing datasets from real-world sources and simulators primarily include scenarios of safe and low-speed driving (below 30 km/h).
In contrast, our DriveCoT includes a substantial portion of high-speed driving data (above 60 km/h), as shown in Fig.~\ref{fig:ego_speed}, posing more challenges for end-to-end driving models.

\textbf{Scenario-level.} The scenario-level statistics, such as scenario types, weather conditions, and time of day, are illustrated in Fig.~\ref{fig:stats_scene}.
The driving scenarios in DriveCoT are categorized into five types according to the hazard conditions encountered by the ego vehicle.
 
\begin{figure}[tb]
  \centering
  \includegraphics[width=0.8\linewidth]{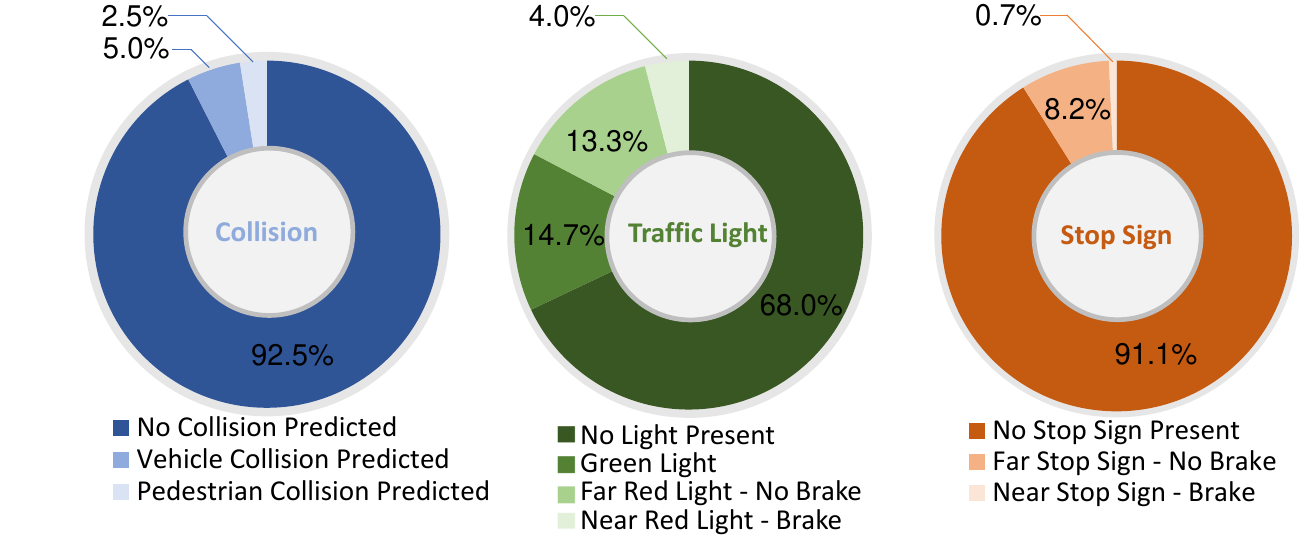}
  % \vspace{-2mm}
  \caption{Sample-level distribution over the chain-of-thought labels in DriveCoT considering potential collision, traffic light status, and stop sign existence.
  }
  \label{fig:stats_CoT}
  % \vspace{-2mm}
\end{figure}

\begin{figure}[tb]
  \centering
  \includegraphics[width=0.8\linewidth]{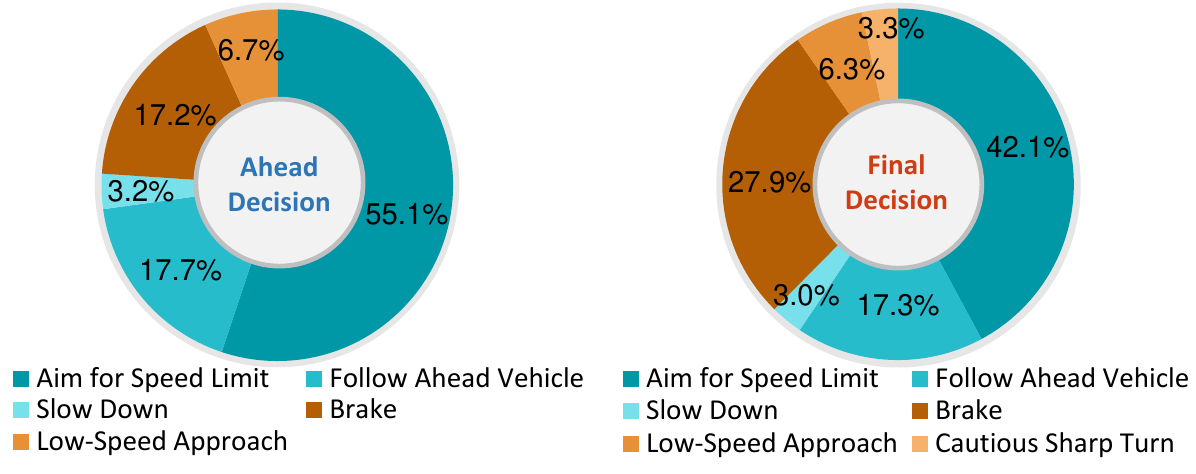}
  \vspace{-2mm}
  \caption{Sample-level distribution over speed decisions. The ahead decision considers the ahead vehicle only, while the final decision also considers other chain-of-thought aspects.
  }
  \label{fig:stats_decision}
  \vspace{-3mm}
\end{figure}

\textbf{Sample-level.} The sample-level statistics regarding the distribution of different CoT aspects and driving decisions are demonstrated in Fig. \ref{fig:stats_CoT} and Fig. \ref{fig:stats_decision}, 
respectively. These figures reveal that DriveCoT contains diverse combinations of CoT processes and final decisions, showcasing its diversity.

\section{DriveCoT-Agent: A Simple CoT Baseline}

\subsection{Network Structure}

\begin{figure}[tb]
  \centering
  \includegraphics[width=1.0\linewidth]{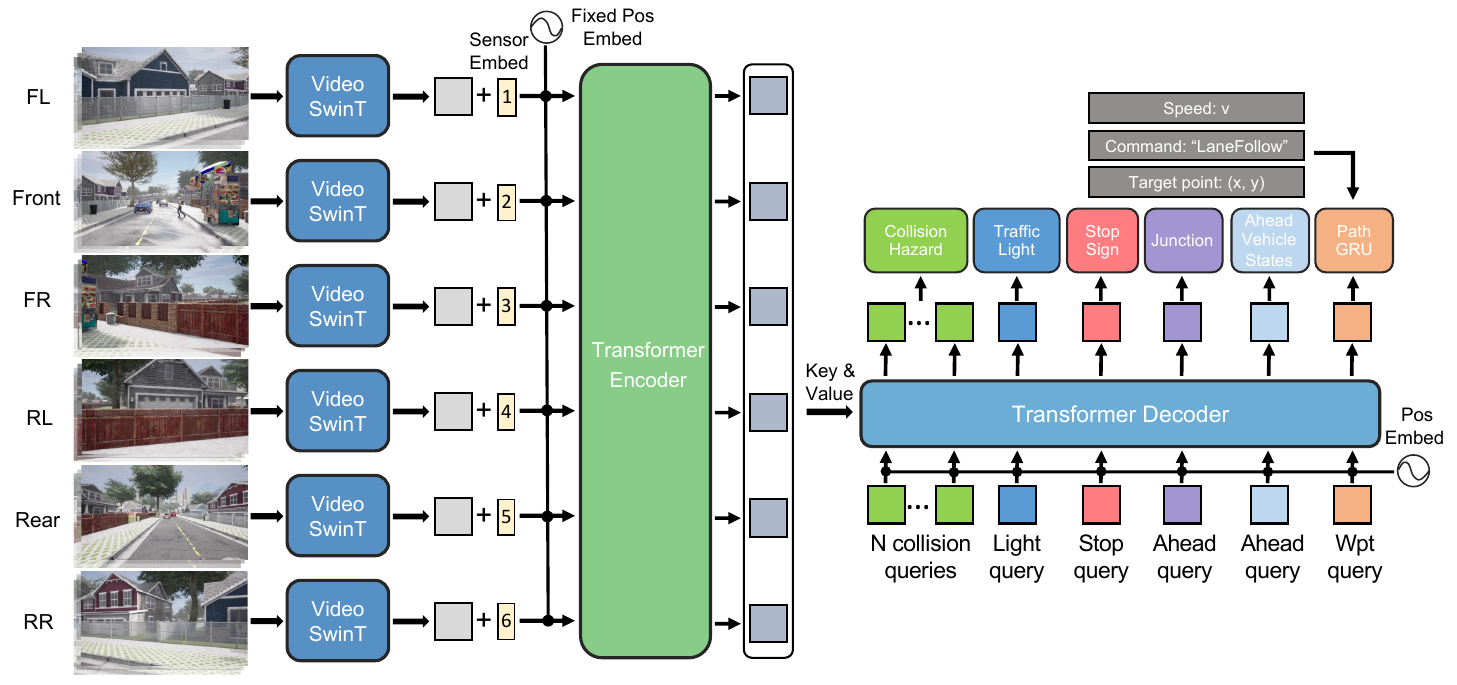}
  \caption{The proposed baseline model DriveCoT-Agent. It takes the multi-view camera videos as inputs and extracts video features for each view via a shared Video SwinTransformer. Then, video tokens of different views are fused via the transformer encoder. For different chain-of-thought driving aspects, we define separate learnable queries for different tasks. This includes collision prediction, recognition of traffic light, stop sign, junction, and prediction of ahead vehicle states. Additionally, the path GRU takes the relevant decoder output with other navigation information to generate the planned waypoints for steering.
  }
  \label{fig:DriveCoT-Agent}
  % \vspace{-2mm}
\end{figure}

We propose a baseline model called DriveCoT-Agent as shown in Fig. \ref{fig:DriveCoT-Agent}.
This model processes video inputs from six surrounding cameras of different views over a specified time period.
It predicts chain-of-thought (CoT) aspects such as potential collisions, traffic light and stop sign hazards, and the relation to the vehicle ahead, so as to determine the target speed and planned waypoints.

\textbf{Input representations.} The inputs include videos captured from six surrounding cameras installed on the ego vehicle, proving the front-left, front, front-right, rear-left, rear, and rear-right views over a recent time period.
Additionally, the path GRU receives information, including the ego vehicle speed, navigation command, and the target point represented as (x, y) coordinates to generate planned waypoints.

\textbf{Output representations.} 
We convert the CoT aspects into a combination of classification and regression tasks. 
Specifically, we use independent prediction heads with linear layers for classifying collision hazards, traffic lights, stop signs, and junction presence, whose possible classes are shown in Fig.~\ref{fig:stats_CoT}.
Besides, the prediction head for ahead vehicle states predicts speed decision considering ahead vehicle only and ahead vehicle's relative speed to ego vehicle. Moreover, the path GRU predicts a path with $L$ waypoints in (x, y) coordinates to guide the ego vehicle's steering direction.

\textbf{Model details.} 
Each video input ${V} \in \mathbb{R}^{T\times3\times H_{0} \times W_{0}}$ are fed into a shared-weight Video-SwinTransformer \cite{videoswin} to extract video feature $\mathbf{f} \in \mathbb{R}^{C\times H\times W}$. In our case, $(H_{0}, W_{0}) = (384, 704)$, $C = 768$, and $(H, W) = (\frac{H_{0}}{32}, \frac{W_{0}}{32})$. The video feature $\mathbf{f}$ is further transformed to lower-dimension $\mathbf{z} \in \mathbb{R}^{d \times H' \times W'}$ via 1x1 convolution and adaptive average pooling. The lower-dimension feature $\mathbf{z}$ is then flattened to tokens, adding fixed sinusoidal positional encoding to retain spatial information and additional learnable sensor embedding to differentiate different views. These tokens from different views are concatenated and then fused by a standard transformer encoder with $K_{enc}$ layers, enabling self-attention across different views.
Further, the fused tokens serve as the keys and values for a transformer decoder with $K_{dec}$ layers, where task-specific learnable query embeddings are used for extracting task-related tokens via an attention mechanism. 
Specifically, we define multiple learnable queries for collision prediction, due to the complexity to model object motion and interactions.
The extracted tokens are directed to different heads to output classification or regression results for various driving aspects through single-layer linear layers.
To predict planned waypoints, following the approaches in \cite{transfuser, hiddenbias, TCP}, we employ a single layer GRU to infer a sequence of $L$ future waypoints $\{\mathbf{w}_{l}\}_{l=1}^{L}$ auto-regressively. It uses the decoder-extracted token, ego speed, navigation command, and target point as inputs.

\textbf{Loss Function.} For the classification tasks, we utilize Focal loss \cite{focalloss} to mitigate the class imbalance issue (see Fig. \ref{fig:stats_CoT}). For the regression tasks, we normalize the ground truths and predictions to fit in relatively small ranges before applying the L1 loss to compute differences. More details can be found in the supplementary material.

\subsection{CoT Process}
\begin{figure}[tb]
  \centering
  \includegraphics[width=1.0\linewidth]{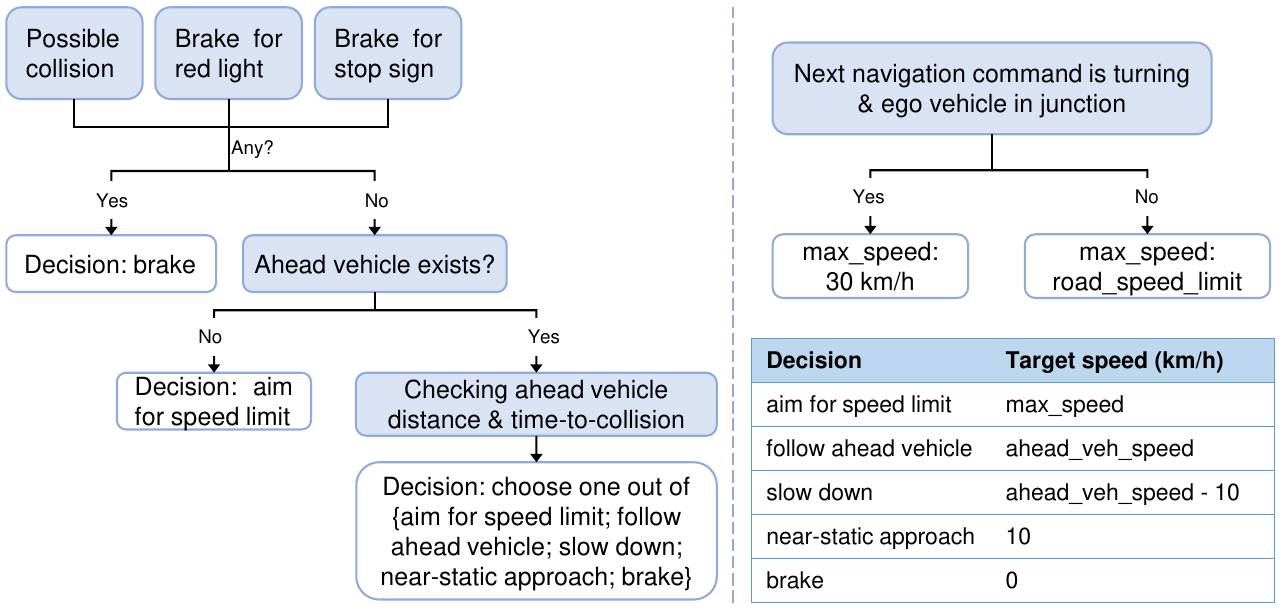}
  \caption{Chain-of-thought process in DriveCoT to obtain the target speed for ego vehicle. It first considers the potential hazards that require emergency brake. If potential hazards are not detected, it further evaluates ego's relation to the ahead vehicle and the road structure. This process decomposes the end-to-end driving into distinctive tasks that provide interpretability and controllability.
  }
  \label{fig:CoT process}
\end{figure}
As shown in Fig.~\ref{fig:CoT process}, we propose a chain-of-thought process to obtain the final target speed based on network outputs. 
Specifically, it first considers the potential hazards that may trigger emergency brake.
If no such hazards are identified, it then evaluates ego's relation to the ahead vehicle and the road structure.
This chain-of-thought process decomposes the end-to-end driving into simple, distinctive tasks, enhancing both interpretability and controllability.
Then, the final target speed and the first waypoint of the predicted planned waypoints will be respectively fed into to a longitudinal PID controller and a lateral steering PID controller for generating low-level control signals, following the methods in~\cite{transfuser, hiddenbias, TCP, interfuser, reasonnet, thinktwice, neat_high_level}.

\section{Experiment}
In the experiment, we investigate the performance of the trained DriveCoT-Agent, including both open-loop evaluation on the validation data of DriveCoT and closed-loop evaluation on CARLA leaderboard 2.0 and Town05Long \cite{transfuser} benchmark.

\textbf{Setup.} The default model parameters include 1-second observation period (3 frames including current frame at 2 Hz setting), video resolution of 384x704, VideoSwinT-tiny backbone, 7x7 adaptive pooling size for each video feature, 12 encoder layers, eight learnable embeddings for collision prediction, and six decoder layers. 
We initialize the model using pre-trained SwinT-tiny weights and conduct training on the DriveCoT training split for 20 epochs.

\subsection{Open-Loop Evaluation}
We conduct open-loop evaluation on the validation split of DriveCoT dataset. Previous end-to-end driving methods such as Transfuser \cite{transfuser} and Interfuser \cite{interfuser} require additional supervision for their specific designs such as depth map or BEV bounding box, which is not supported in DriveCoT. As a result, we do not re-train the models and only evaluate their zero-shot performance on DriveCoT. Additionally, these methods only consider driving safety, thus can only extract the binary decisions (normal drive or brake). 
For DriveCoT-Agent, we use argmax to determine the speed decision. As for the metrics, we calculate the F1 score of classification for each gt speed decision class.
For the predicted waypoints, we calculate the angle derived from the chosen waypoint and compare it with the ground truth. If the angle difference is within 2 degrees, it is considered accurate prediction.

As illustrated in Table \ref{tab:open-loop evaluation}, our proposed DriveCoT-Agent can predict more detailed and precise decisions compared to existing methods and can also provide reasons. We also compare DriveCoT-Agent with a variant that shares the similar network and directly predict the final speed decision. DriveCoT-Agent performs better on all aspects, showing the benefits brought by chain-of-thought process.

\begin{table}[t!]
    \caption{Open-loop evaluation on DriveCoT dataset val split. Previous methods can only extract binary speed decisions (normal drive or brake). Compared to previous methods, the proposed DriveCoT-Agent can predict more precise and detailed speed decisions and steering waypoints.}
    \vspace{-1mm}
    \begin{center}
    \renewcommand{\tabcolsep}{1mm}
        \resizebox{1.0\linewidth}{!}{
        \begin{tabular}{l|c c c c c c|c c c}
            \toprule[0.3mm]
            
            % \multirow{2}{*}{Dataset} & \multicolumn{6}{c|}{Speed (F1)} &  \multicolumn{3}{c|}{Path (accuracy)}\\
            \multirow{2}{*}{Method} & \multicolumn{6}{c|}{Speed (F1 $\uparrow$)} & \multicolumn{3}{c}{Path (accuracy $\uparrow$ \%)} \\
             & SpeedLimit & FollowAhead & SlowDown & SlowApproach & CautiousTurn & Brake & Straight & Turn & LaneChange\\
            \midrule
            Transfuser \cite{transfuser} & - & - & - & - & - & 0.10 & 60.6 & 40.1 & 31.1\\
            TCP \cite{TCP} & - & - & - & - & - & 0.21 & 63.1 & 42.5 & 29.0\\
            Interfuser \cite{interfuser} & - & - & - & - & - & 0.35 & 62.6 & 38.1 & 27.3\\
            % \textbf{DriveCoT-Agent (direct)} & xx & xx.x & xx.x & xx.x & xx.x & xx.x & xx.x & xx.x & xx.x\\
            \midrule
            direct decision & 0.61 & 0.59 & 0.32 & 0.50 & 0.31 & 0.41 & 84.1 & 74.2 & 75.1\\
            \textbf{DriveCoT-Agent} & \textbf{0.87}  & \textbf{0.81} & \textbf{0.75} & \textbf{0.72} & \textbf{0.83} & \textbf{0.84} & \textbf{87.2} & \textbf{76.1} & \textbf{79.8}\\
            \bottomrule[0.3mm]
        \end{tabular}
        }
    \label{tab:open-loop evaluation}
    \end{center}
\end{table}
% \vspace{-15mm}

\subsection{Ablation Study} 
To determine the optimal model configurations of DriveCoT-Agent, we conduct ablated experiments on model parameters.

\textbf{Pooling size, encoder layers, and collision embeddings.}
As demonstrated in Table \ref{tab:ablation_parameter}, our chosen configurations can achieve optimal performance. Specifically, path accuracy is relatively stable with different model parameters since predicting the disentangled waypoints to indicate direction is not a very complex task. Besides, the inclusion of transformer encoder layers and multiple learnable query embeddings for collision prediction improves the performance of speed decisions.

\begin{table}[t!]
    \caption{Ablations on model parameters of DriveCoT-Agent, including pooling size, encoder layers, and collision embeddings. The chosen optimal configs are highlighted.}
    \vspace{-2mm}
    \begin{center}
    \renewcommand{\tabcolsep}{1mm}
        \resizebox{0.6\linewidth}{!}{
        \begin{tabular}{l|c c|c c c}
            \toprule[0.3mm]
            
            % \multirow{2}{*}{Dataset} & \multicolumn{6}{c|}{Speed (F1)} &  \multicolumn{3}{c|}{Path (accuracy)}\\
            \multirow{2}{*}{Configurations} & \multicolumn{2}{c|}{Speed (F1 $\uparrow$)} & \multicolumn{3}{c}{Path (accuracy $\uparrow$ \%)} \\
             & Non-brake & Brake & Straight & Turn & LaneChange\\
            \midrule
            1x1 pooling & 0.51 & 0.30 & 40.2 & 35.3 & 30.1\\
            3x3 pooling & 0.72 & 0.78 & 86.3 & \textbf{77.7} & 76.2\\
            \textbf{7x7 pooling} & \textbf{0.82} & \textbf{0.84} & 87.2 & 76.1 & \textbf{79.8}\\
            9x9 pooling & 0.77 & 0.80 & \textbf{88.1} & 76.5 & 77.2\\
            \midrule
            0 encoder layers & 0.61 & 0.59 & 84.2 & 75.1 & 74.3\\
            3 encoder layers & 0.75 & 0.68 & 85.1 & 73.2 & 76.1\\
            6 encoder layers & 0.73 & 0.62 & 82.3 & 74.5 & 75.6\\
            \textbf{12 encoder layers} & \textbf{0.82} & \textbf{0.84} & \textbf{87.2} & \textbf{76.8} & \textbf{79.8}\\
            15 encoder layers & 0.71 & 0.65 & 83.5 & 72.2 & 73.7\\
            \midrule
            1 collision embeds & 0.71 & 0.51 & 84.5 & 74.7 & 73.1\\
            3 collision embeds & 0.74 & 0.63 & 85.1 & 75.2 & 74.8\\
            6 collision embeds & 0.73 & 0.75 & \textbf{88.3} & 75.9 & 77.6\\
            \textbf{8 collision embeds} & \textbf{0.82} & \textbf{0.84} & 87.2 & \textbf{76.1} & \textbf{79.8}\\
            10 collision embed & 0.71 & 0.72 & 87.9 & 75.4 & 76.3\\
            
            \bottomrule[0.3mm]
        \end{tabular}
        }
    \label{tab:ablation_parameter}
    \end{center}
\end{table}

\textbf{Input horizon.}
We further investigate the impact of the input horizon length. As shown in Table \ref{tab:ablation_horizon}, DriveCoT-Agent's model variant with only single frame inputs performs much worse than the models with video inputs. For instance, the model with single frame inputs only achieves an F1 score of 0.25 on brake decision, while the model with 0.5-second video inputs (2 frames at 2 Hz) gets an F1 score of 0.69. 

Moreover, increasing the input horizon from 0.5 s to 1 s can largely improve the speed decision and path accuracy. This comes from the fact that the acceleration of objects can be derived from 3 frames such that the F1 score for brake decision becomes much higher. When further increasing the input horizon, the performance gain becomes marginal and even worse than shorter horizons. This might stem from the simple design for video feature extraction in our proposed DriveCoT-Agent, indicating a direction for future study. 

The increasing input horizon may result in slower inference. We calculate the inference speed on an RTX 3090 GPU and average the time spent on all evaluated data in Table \ref{tab:ablation_horizon}. Our chosen 1-s configuration can achieve both satisfactory performance and a fast inference speed of 23.1 FPS.

\begin{table}[t!]
    \caption{Ablations on input horizon of DriveCoT-Agent. Our chosen 1-second configuration achieves both satisfactory performance and fast inference speed.}
    \begin{center}
    \renewcommand{\tabcolsep}{1mm}
        \resizebox{0.65\linewidth}{!}{
        \begin{tabular}{l|c c|c c c|c}
            \toprule[0.3mm]
            \multirow{2}{*}{Input Horizon} & \multicolumn{2}{c|}{Speed (F1 $\uparrow$)} & \multicolumn{3}{c|}{Path (accuracy $\uparrow$ \%)} & \multirow{2}{*}{FPS}\\
             & Non-brake & Brake & Straight & Turn & LaneChange &\\
            \midrule
            single frame & 0.40 & 0.25 & 82.1 & 72.3 & 74.1 & 37.1\\
            0.5 seconds & 0.65 & 0.69 & 84.1 & 73.7 & 75.2 & 26.9\\
            \textbf{1 seconds} & \textbf{0.82} & 0.84 & \textbf{87.2} & \textbf{76.1} & \textbf{79.8} & 23.1\\
            2 seconds & 0.80 & \textbf{0.85} & 86.1 & 75.7 & 77.2 & 21.1\\
            3 seconds & 0.76 & 0.79 & 83.2 & 72.4 & 76.0 & 18.6\\
            \bottomrule[0.3mm]
        \end{tabular}
        }
    % \vspace{-2mm}
    \label{tab:ablation_horizon}
    \end{center}
\end{table}

\subsection{Closed-Loop Evaluation}
 Closed-loop evaluation is more challenging since the system's actions directly impact the state of the environment compared to the fixed recorded evaluation data in an open-loop setting. We evaluate the trained DriveCoT-Agent in the CARLA simulator to apply the model decisions to the ego vehicle in the simulator. 
 
 The original trained model sometimes predicts potential collision with a near object, thus stuck at a place while the object is static and has no risk of colliding. We use a simple data augmentation strategy to finetune the trained model to solve this issue. We select out the data with its CoT label of potential collision being true (see Fig. \ref{fig:stats_CoT}) and create new data that has the same images as the current frame over the observation period, e.g. 1s to simulate a static video where none of the objects move. In addition, we alter the augmented data's CoT label of potential collision to be false since all objects are static such that there is no potential for collision. We then mix the augmented data (\textasciitilde 1.8k) with the original training data (\textasciitilde 23.8k) to finetune the model for five epochs such that forcing the model to consider the object motion embedded in the video input. 
 Our finetuned model is then evaluated on a widely used Town05Long Benchmark \cite{transfuser} and leaderboard 2.0 with unseen validation routes.
 
 As shown in Table \ref{tab:closed_loop_eval_town05}, DriveCoT-Agent surpasses the previous methods on the Town05Long benchmark with higher Driving Score, Route Completion, and Infraction Score. When evaluating the more challenging leaderboard 2.0 scenarios, we choose some officially provided routes that were not included when collecting the DriveCoT dataset and further split these long routes into 2 km each to align with the Town05Long benchmark. As illustrated in Table \ref{tab:closed_loop_eval_2.0}, DriveCoT-Agent achieves satisfactory driving performance, while previous methods all perform poorly. 

\begin{table}[t!]
    \caption{Closed-loop evaluation on Town05Long and leaderboard 2.0 with unseen routes. DS denotes driving Score, RC denotes Route Completion, and IS denotes Infraction Score.}
    \centering
    \captionsetup[sub]{font=footnotesize}
    \begin{subtable}{0.4\linewidth}
        \centering
            \resizebox{0.75\linewidth}{!}{
            \begin{tabular}{l|c c c}
                \toprule[0.3mm]
                Method & DS $\uparrow$ & RC $\uparrow$ & IS $\uparrow$ \\
                \midrule
                Transfuser \cite{transfuser} & 31.0 & 47.5 & 0.77 \\
                TCP \cite{TCP} & 57.2 & 80.4 & 0.73 \\
                Interfuser \cite{interfuser} & 68.3 & 95.0 & 0.72 \\
                \textbf{DriveCoT-Agent} & \textbf{73.6} & \textbf{96.8} & \textbf{0.76} \\
                \bottomrule[0.3mm]
            \end{tabular}
            }
        \caption{Town05 Long}
        \label{tab:closed_loop_eval_town05}
    \end{subtable}
    % \hfill
    \hspace{10pt}
    \begin{subtable}{0.4\linewidth}
        \centering
            \resizebox{0.75\linewidth}{!}{
            \begin{tabular}{l|c c c}
                \toprule[0.3mm]
                Method & DS $\uparrow$ & RC $\uparrow$ & IS $\uparrow$ \\
                \midrule
                Transfuser \cite{transfuser} & 1.3 & 10.2 & 0.13 \\
                TCP \cite{TCP} & 3.1 & 13.5 & 0.23 \\
                Interfuser \cite{interfuser} & 2.8 & 15.6 & 0.18 \\
                \textbf{DriveCoT-Agent} & \textbf{51.9} & \textbf{77.5} & \textbf{0.67} \\
                \bottomrule[0.3mm]
            \end{tabular}
            }
        \caption{leaderboard 2.0 unseen routes}
        \label{tab:closed_loop_eval_2.0}
    \end{subtable}
    \label{tab:closed_loop_eval}
\end{table}
\vspace{-2mm}

\subsection{Qualitative Results \& Visualization}
\begin{figure}[tb]
  \centering
  \includegraphics[width=0.95\linewidth]{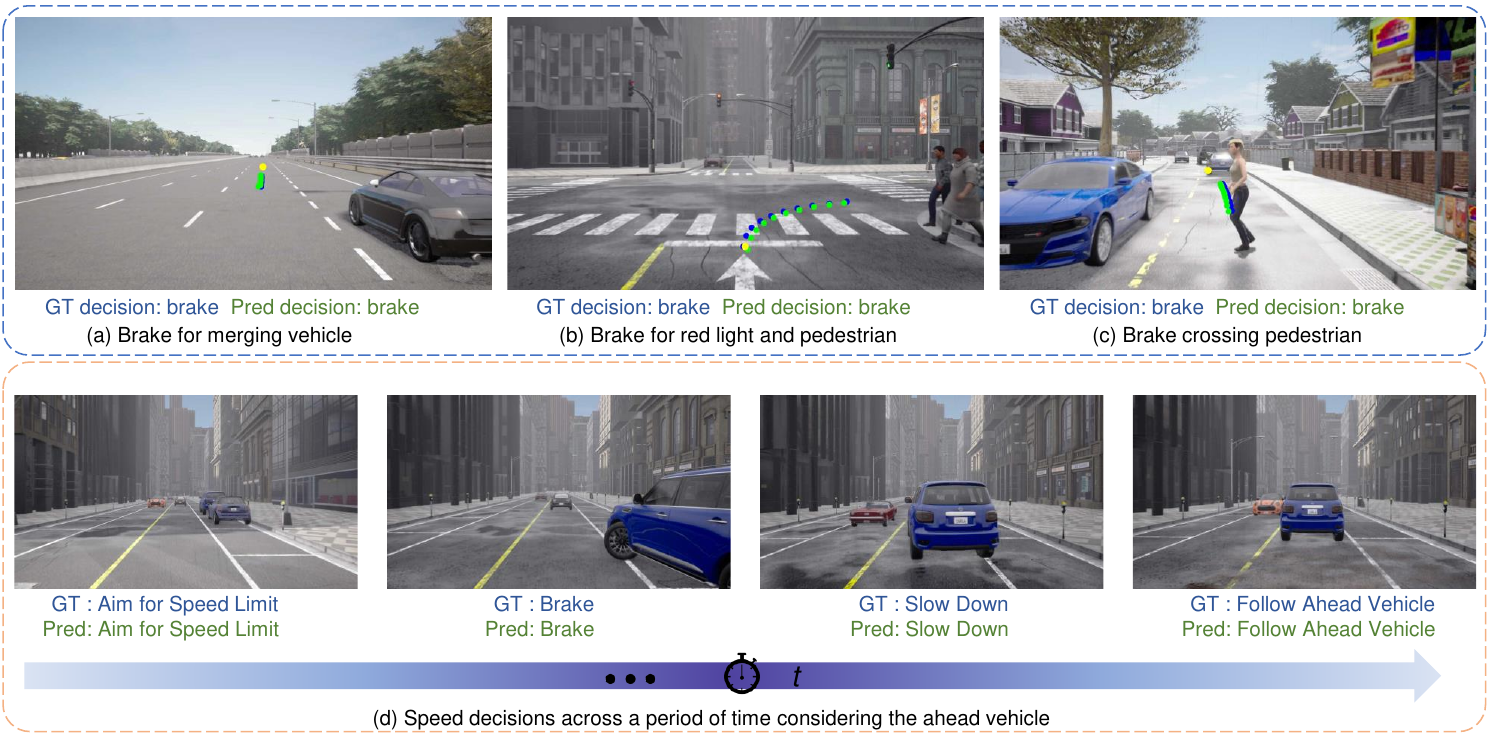}
  \caption{Qualitative results of DriveCoT-Agent. It correctly brakes for (a) the lane-merging vehicle, (b) the red traffic light and pedestrian, and (c) the crossing pedestrian in the middle of the road. The yellow dot in the image is the target point, which indicates the direction, while the blue and green dots represent the ground truth and predicted future waypoints. In (d), DriveCoT-Agent generates appropriate speed decisions concerning the ahead vehicle based on the distance and time-to-collision information embedded in the video input.
  }
  \label{fig:qualitative_results}
\end{figure}
% \vspace{-2mm}

\begin{figure}[tb]
  \centering
  \includegraphics[width=0.95\linewidth]{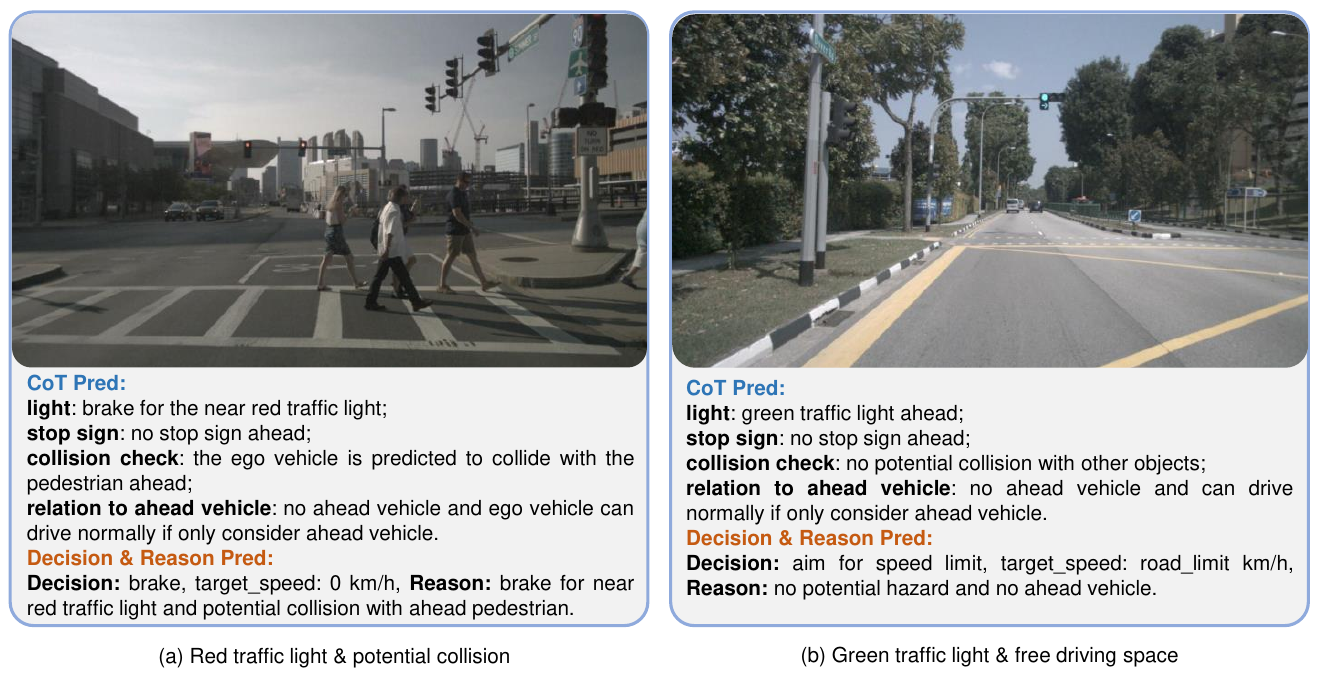}
  \caption{Zero-shot performance of DriveCoT-Agent on real-world dataset nuScenes. It can (a) recognize the red light and potential collision with pedestrian, thus brake and (b) identify the green light ahead and drive normally due to the absence of ahead vehicle at its proximity.
  }
  \label{fig:qualitative_results_nuscenes}
\end{figure}

This section provides some qualitative results to visualize DriveCoT-Agent's driving and reasoning ability for different scenarios.

\textbf{Visualization on DriveCoT data.} Fig. \ref{fig:qualitative_results} presents the qualitative results of DriveCoT-Agent. 
In Fig. \ref{fig:qualitative_results} (a), it detects the ahead merging vehicle and correctly predicts potential collision and thus brakes. In Fig. \ref{fig:qualitative_results} (b), it identifies the red traffic light and the potential collision with pedestrian and thus brakes. In Fig. \ref{fig:qualitative_results} (c), it recognizes the crossing pedestrian in the middle of the road, predicts potential collision, and thus brakes. Moreover, as shown in Fig. \ref{fig:qualitative_results} (d), DriveCoT-Agent generates proper speed decisions concerning the ahead vehicle across a time period based on the distance and time-to-collision.

\textbf{Performance on real-world data.}
To validate the generalizability of DriveCoT-Agent to real-world data. We apply DriveCoT-Agent on the scenes dataset in a zero-shot manner. Interestingly, it can already generate reasonable decisions and reasons, as demonstrated in Fig. \ref{fig:qualitative_results_nuscenes}, demonstrating the usefulness of the proposed DriveCoT data and the trained DriveCoT-Agent model. For example, in Fig. \ref{fig:qualitative_results_nuscenes} (a), DriveCoT-Agent recognizes the red traffic light ahead and predicts potential collisions with pedestrians, thus deciding to brake. In Fig. \ref{fig:qualitative_results_nuscenes} (b), DriveCoT-Agent identifies the green traffic light and free space ahead, thus decide to drive normally and aim for the road speed limit.

\section{Conclusion}
In this paper, we have proposed a new dataset, DriveCoT, which is designed for end-to-end driving with chain-of-thought labels on different driving aspects and final driving decisions. The dataset contains the thinking process labels when making driving decisions, aiming to stimulate further research on driving with understanding and serve as an effective evaluation benchmark. Additionally, we propose a baseline model, DriveCoT-Agent, that generates predictions on different aspects to offer interpretability and obtains the final decision through a chain-of-thought process. Notably, DriveCoT-Agent exhibits better performance compared to existing methods in both open-loop and closed-loop testing, showing the benefits of integrating chain-of-thought with end-to-end driving.

\clearpage  % TODO REVIEW/FINAL: This \clearpage needs to be removed from both review and camera-ready versions.

\bibliographystyle{splncs04}
\bibliography{main}

\newpage
\appendix
% \chapter{appendix}
\section{Expert Policy}
For the expert policy that we use to control the ego vehicle during data collection, we mainly follow the method proposed in \cite{carlaexpert} with some modifications. Specifically, these changes include the dynamic braking distance, the varying collision prediction horizon, and additional consideration of relation to the ahead vehicle to support high-speed driving and traffic-aligned behavior. In the expert policy, we disentangle the target speed with the planned waypoints where the former is determined by various driving aspects and the latter is pure based on the map topology and the ego vehicle speed. The simulation \& control frame rate are both 20 FPS, while the data collection frequency is set to be 2 Hz to avoid redundant data.

\subsection{Target Speed}
As shown in Fig. \ref{fig:CoT process}, the driving aspects that the expert policy considers when determining the target speed include the checking of \textit{potential hazards over red traffic light, stop sign, and collision}, \textit{relation to the ahead vehicle}, and 
\textit{whether approaching sharp turns}.

\textbf{Light \& stop sign hazard.} In CARLA, we can use the provided API to get the influenced road and lane IDs for each traffic light and stop sign. Then, based on the ego vehicle's current and future planned driving lane, we can filter out the red traffic lights and stop signs that could potentially impact the ego vehicle. We can obtain the trigger volumes of these filtered lights and stop signs, which are represented as bounding boxes. We then put a safety-checking bounding box in front of the ego vehicle to initiate braking when there is an overlap with the hazard trigger volumes. We use CARLA version 0.9.14 in this paper, where the vehicle dynamics have been tuned to be more realistic compared to the previous versions. This change has a noticeable impact on the braking distance, where the vehicle could stop unreasonably quickly in previous versions and now requires a longer and dynamic distance to a full stop. 
Considering the varying braking distance required for a full stop from diverse ego vehicle speeds, we assign the safety-checking bounding box with a dynamic length defined as the safety distance to check potential hazards and adapt to high-speed driving. Specifically, we define the safety distance $d\_\text{safety}$ as:

{\small
\[
d\_\text{safety} = \begin{cases}
3 & \text{if } \mathit{v} < 30\ \text{km/h}\\
\dfrac{(\mathit{v} / 3.6)^2}{2 * | {\mathit{a}\text{\_brake\_max}} |} - 4
& \text{otherwise}
\end{cases}
\]
}
where $v$ is the ego vehicle speed in km/h and $a$\_brake\_max is the maximum deceleration for brake, which we test and choose to be -5 m/s*2. In this equation, we choose a fixed safety distance for an ego speed under 30 km/h and a dynamic distance for a higher ego speed. For example, in our case, the ego vehicle with 72 km/h would require a safety distance of 36 meters.

\begin{figure}[tb]
  \centering
  \includegraphics[width=1.0\linewidth]{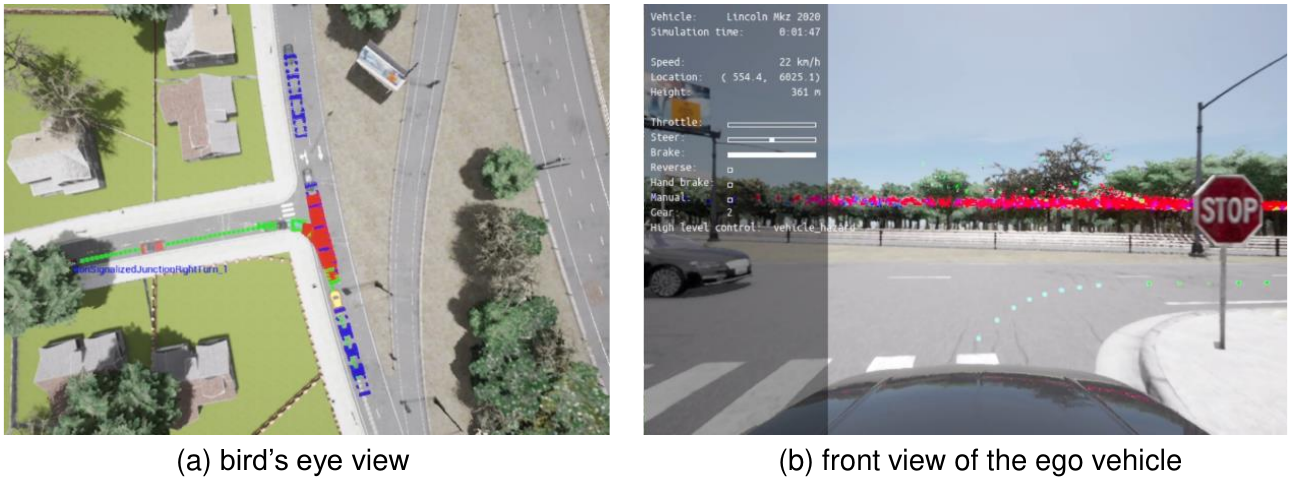}
  \caption{
An example of expert policy checking potential collision with ego vehicle. The ego vehicle aims to turn right while encountering incoming vehicles that go straight. In (a), the bird's eye view is centered on the ego vehicle. The dense, small dots in green, blue, and gray are the predefined route waypoints with semantic meanings. Besides, the predicted future bounding boxes for the ego vehicle and other vehicles are normally drawn as green and blue, while the overlapped boxes are drawn as red. As a result, in (b), the expert policy chooses to brake now to avoid a potential collision.}
  \label{fig:collision_check}
\end{figure}

\textbf{Collision hazard.} To predict the ego vehicle's potential collision with other vehicles and pedestrians, we need to predict their future trajectories. Following \cite{carlaexpert}, we use a bicycle kinematics model to predict their future positions and orientations iteratively. The actions of other vehicles and pedestrians are assumed to remain the same as the current step. In contrast, the ego vehicle's future actions are dynamically changed by PID controllers to follow the planned waypoints virtually. For the future horizon length we use to check collision, we set it to two seconds when ego speed is lower than 80 km/h and otherwise three seconds. Under our 20 FPS simulation frame rate, we need to check whether the ego vehicle's future trajectory has overlapped with other agents for the subsequent 40 or 60 frames. In Fig. \ref{fig:collision_check} (a), we plot the future bounding boxes of the ego vehicle (center of Fig. \ref{fig:collision_check} (a)) and other vehicles over the future horizon. We plot the bounding boxes of the ego vehicle as green and the other vehicles as blue and change the color to red if overlap is detected. Moreover, when we detect potential collisions with the same agent for more than five consecutive frames, we consider that agent as a dangerous agent and change the collision criteria to that agent as three meters away instead of overlapping to avoid close contact.

\textbf{Relation to ahead vehicle.} To generate more appropriate and traffic-aligned driving behaviors, we further consider the ego vehicle's relation to the ahead vehicle in the same driving lane. Based on the distance and time-to-collision to the ahead vehicle, the ego vehicle chooses one action out of \{\textit{aim for speed limit; follow ahead vehicle; slow down; near-static approach; brake}\} when ahead vehicle exists. Specifically, when the distance to the ahead vehicle is less than 5 meters or a distance of less than 10 meters \& the speed difference is less than 3 m/s, we use the distance to choose actions and otherwise leverage the time-to-collision. Taking the distance factor as an example, as the ego vehicle gets closer to the ahead vehicle, the appropriate action gradually changes from \textit{aim for speed limit}, to \textit{follow ahead vehicle}, to \textit{slow down}, and to \textit{brake}. In addition, we enlarge the distance interval for the current decision, such as \textit{follow ahead vehicle}, to achieve a damping effect and avoid frequent decision switches. The specific target speeds of different decisions are listed in Fig. \ref{fig:CoT process} and can be altered flexibly. Especially when the ego target speed and the ahead vehicle speed are both slower than 5 km/h, we set the decision to be \textit{near-static approach} with 10 km/h target speed.
Note that the ahead\_decision will only be chosen as the final\_decision when there are no potential hazards. 

\textbf{Approach sharp turns.} Approaching sharp turns at high speed can potentially cause instability of the ego vehicle. As a result, we set the maximum target speed to 30 km/h when the following navigation command is turning and the ego vehicle is currently inside of a junction.

\subsection{Planned Waypoints}
Given the map topology and the predefined sparse waypoints in a route, we can obtain a dense set of waypoints with fixed distance as shown in Fig. \ref{fig:collision_check} to complete the route. Following \cite{hiddenbias}, we choose a sequence of ten future waypoints $\{\mathbf{w}_{l}\}_{l=1}^{L}$ with fixed distance-gap of 1 meters. Unlike previous works, we choose the planned waypoints more distant from the ego vehicle as the ego vehicle speed increases to avoid oscillating, which also aligns with human driving choices. The distance to the first waypoint $d\_\text{wpt}$ is defined as:

{\small
\[
d\_\text{wpt} = \begin{cases}
4 & \text{if } \mathit{v} < 20\ \text{km/h}\\
% \dfrac{(\mathit{v} / 3.6)^2}{2 * | {\mathit{a}\text{\_brake\_max}} |} - 4
0.5*\dfrac{\mathit{v}}{3.6} + 2
& \text{otherwise}
\end{cases}
\]
}
where $\mathit{v}$ is the ego vehicle speed in km/h. For execution, only the first waypoint will be utilized to guide the ego vehicle's steering, and the rest of the waypoints are to provide extra supervision, which has proven to be beneficial for the driving performance in \cite{transfuser, hiddenbias, interfuser}.

\subsection{PID Controller}
The chosen target speed and future waypoint are respectively sent to a longitudinal PID controller for acceleration \& brake and a lateral PID controller for steering, following previous works \cite{transfuser, hiddenbias, TCP, interfuser, reasonnet, thinktwice, neat_high_level}. The tuned PID controllers are supposed to adapt to both low-speed and high-speed driving situations.
The tuned longitudinal PID controller takes ego speed and target speed in km/h as inputs and sets K\_P, K\_I, K\_D to 0.3, 0.05, 0 with a 20-frame buffer to calculate the averaged integral term. For the tuned lateral PID controller, it takes the angle difference in radians between the ego vehicle heading and the vector pointing the chosen future waypoint as input and sets K\_P, K\_I, K\_D to 0.8, 0.3, 0 with a 10-frame buffer to calculate the averaged integral term. 

\section{Driving Scenarios}
During the data collection of the DriveCoT dataset and the closed-loop testing for the trained DriveCoT-agent, we utilize the CARLA leaderboard 2.0 framework \cite{leaderboard2.0} to run the simulation. The dataset collection and closed-loop testing routes are chosen from the predefined routes officially provided by the CARLA team in the latest large map Town12 without overlap. For the involved driving scenarios, we adapt most of the designed scenarios in Leaderboard 2.0 and ignore those that force an overtaking behavior, such as overtaking the obstacles that block the road. The overtaking scenarios require overly complicated expert policy and model design to have a memory for past decisions and reasons, which we plan to extend in our future work.

\section{Dataset Annotations}
\begin{figure}[tb]
  \centering
  \includegraphics[width=0.95\linewidth]{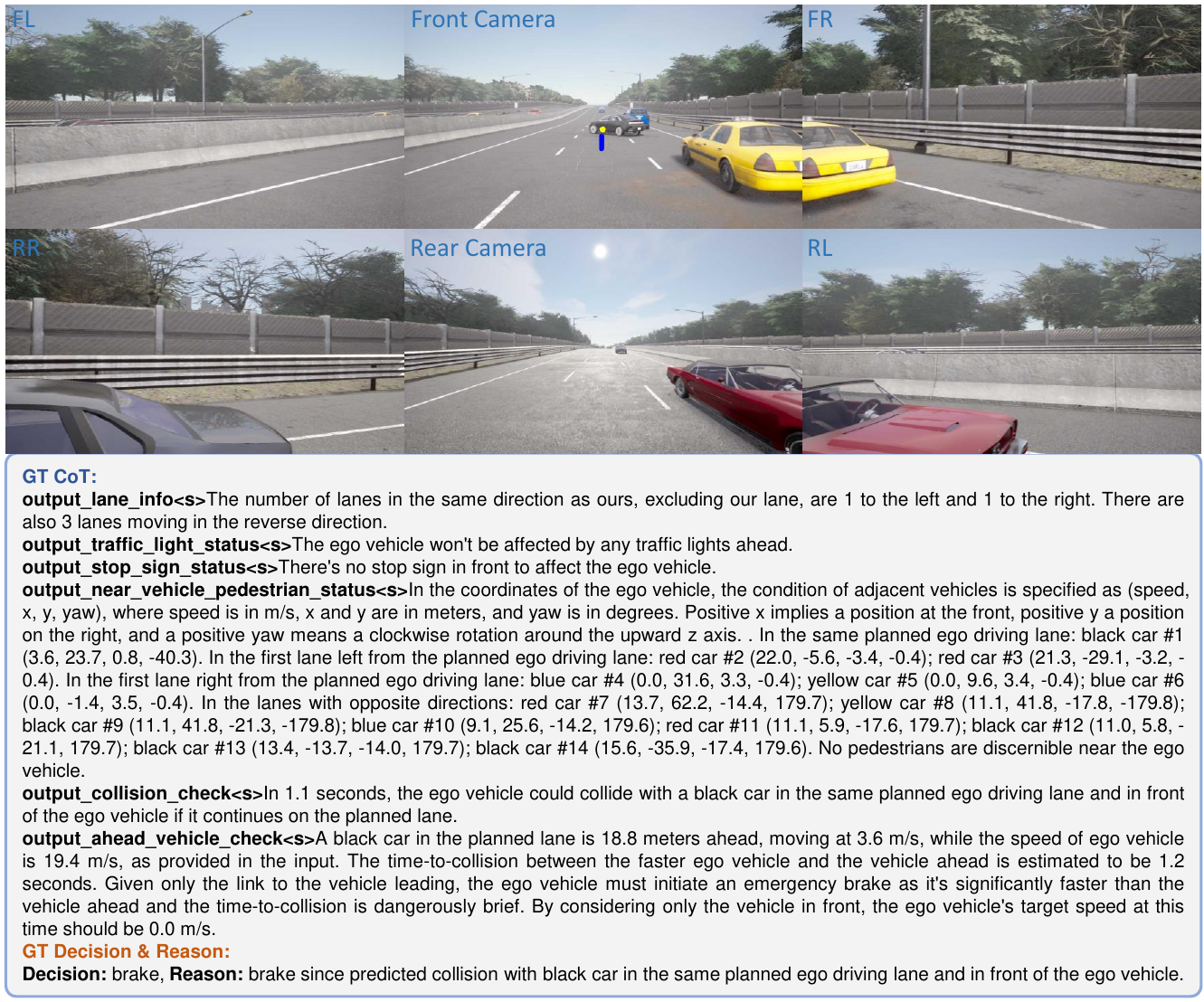}
  \caption{An example of the text-form annotations in DriveCoT dataset. Besides the chain-of-thought labels shown in the main paper, we also provide perception labels to indicate other agents' precise locations and orientations. In this case, the ego vehicle is projected to collide with the ahead black car that slowly cuts in from a blocked driving lane. Thus, the expert policy decides to brake at the moment. Additionally, the yellow dot in the image is the target point, which indicates the direction, while the blue dots represent the ground truth waypoints to guide steering.
  }
  \label{fig:supple_text_annotation_full}
\end{figure}

We provide an example of the detailed annotations of DriveCoT dataset in Fig. \ref{fig:supple_text_annotation_full}. Besides the chain-of-thought labels shown in the main paper, we also provide perception labels to indicate other agents’
precise locations and orientations. In this case, the expert policy predict a potential collision with the ahead black car, thus brake.
\vspace{-3mm}

\section{Implementation Details}
\subsection{Model Details}
The video inputs of each view have the original 1600x900 resolution. During preprocessing, the captured images from the same video are resized and cropped with identical randomly chosen parameters to offer data diversity and ensure temporal consistency. The processed videos have a resolution of 704x384. The processed videos are then independently fed into the pre-trained VideoSwinTransformer-tiny backbone to extract the per-view video features. The extracted video features are then collapsed into 1D tokens and added with learnable sensor embeddings and fixed 2D sinusoidal positional encodings to differentiate views and keep spatial information. Tokens of different views are concatenated and fed into a transformer encoder with 12 encoder layers to fuse the information. After that, the fused tokens serve as the key and value for the transformer decoder, which consists of six decoder layers. Then, separate learnable query embeddings are defined for different subtasks to extract task-relevant features from the transformer decoder, which are fed into corresponding prediction heads for various classification and regression tasks. As illustrated in Fig. \ref{fig:DriveCoT-Agent}, the network outputs include the classification of collision hazard, light hazard, stop sign hazard, junction status, ahead vehicle existence, and decision considering ahead vehicle only, and regression of the ego speed, the relative speed of the ahead vehicle, and the future planned 2D waypoints. Fig. \ref{fig:stats_CoT} shows the possible classification categories of different CoT aspects.

\subsection{Loss Function \& Training}
\textbf{Classification tasks.} For the numerous classification tasks we list above, we leverage Focal loss \cite{focalloss} to tackle the class imbalance issues. We set alpha and gamma to 0.95 and 2 to emphasize the positive samples and hard examples.

\textbf{Regression of the ego speed \& the relative ahead vehicle speed.} We normalize the ego speed and the relative ahead vehicle speed by dividing them by 30 and 10 km/h, respectively. Then, L1 loss is applied between the normalized predictions and ground truth labels.

\textbf{Regression of the future planned 2D waypoints.} The DriveCoT-agent predicts a series of 2D waypoints (x,y) in local coordinates. We design a customized loss function, WaypointL1Loss, which prioritizes the accuracy of predictions for waypoints that are in closer future while also giving more importance to the lateral dimension over the longitudinal dimension in the predicted waypoints. This is achieved through a weighting matrix that assigns greater weights to the immediate future waypoints and the lateral errors, thereby focusing the model's learning more on the short-term trajectory precision and lateral positioning accuracy.

\textbf{Coefficients of different loss terms.} We assign a coefficient of 0.1 to the  WaypointL1Loss and 1.0 for all the other loss terms. 

% \section{More Examples of the Proposed DriveCoT Dataset}
\section{More Visualizations}
\begin{figure}[tb]
  \centering
  \includegraphics[width=0.95\linewidth]{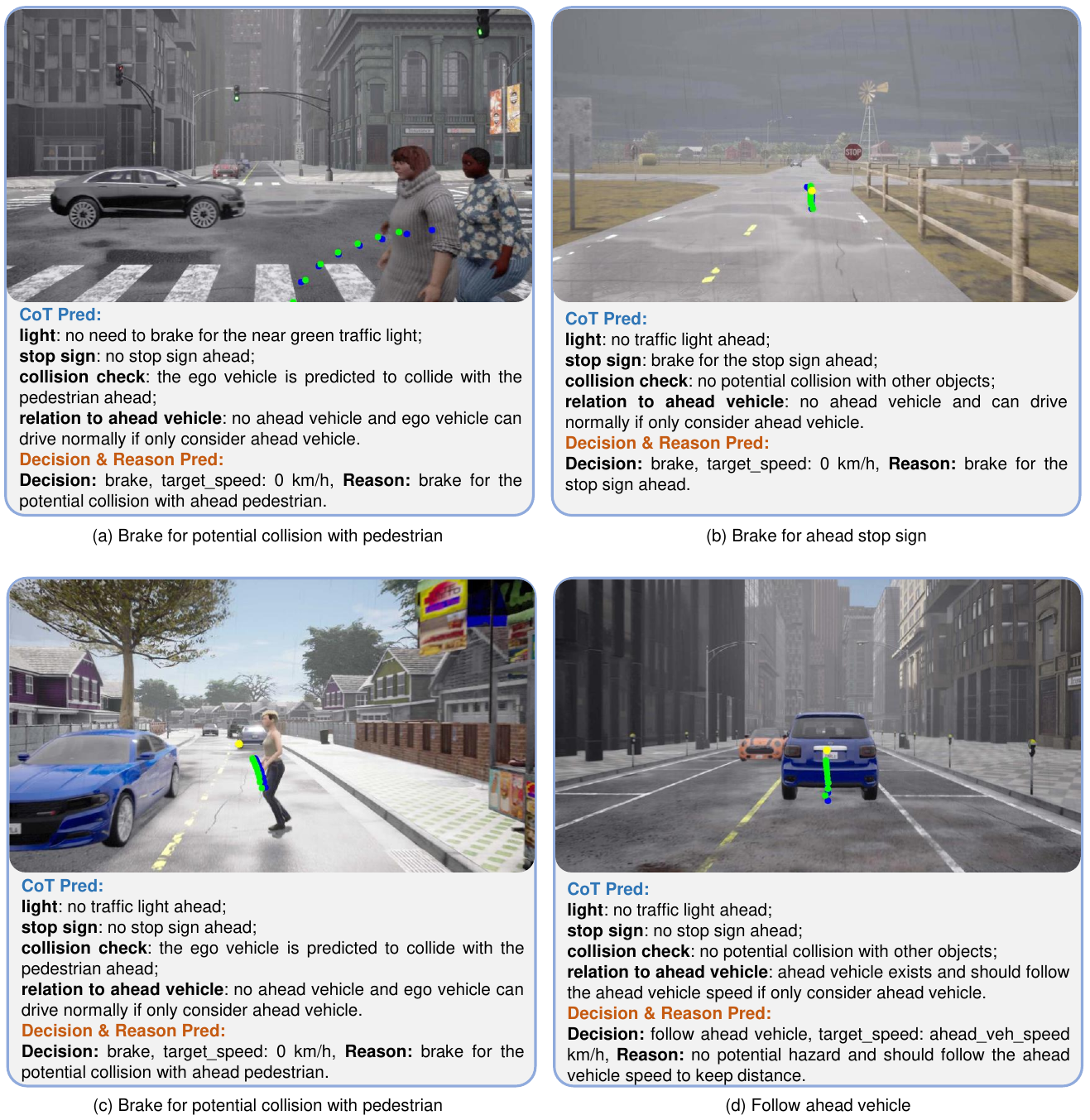}
  \caption{Additional qualitative results of our trained DriveCoT-agent on the validation split of DriveCoT dataset. The chain-of-thought processes used to obtain the target speeds are shown below the images . The yellow dot in the image is the target point, which indicates the direction, while the blue and green dots represent the ground truth and predicted future waypoints.
  }
  \label{fig:supple_qualitative_results_carla}
\end{figure}

In Fig. \ref{fig:supple_qualitative_results_carla}, we provide more visualization results of the trained DriveCoT-Agent on the validation split of DriveCoT dataset

\end{document}